\pdfoutput=1
\documentclass[lettersize,journal]{IEEEtran}

\usepackage{array}
\newcolumntype{M}[1]{>{\centering\arraybackslash}m{#1}}
\newcolumntype{N}{@{}m{0pt}@{}}
\usepackage{boldline}

\usepackage{cite}
\usepackage{amsmath}
\usepackage{amssymb}
\usepackage{amsthm}
\usepackage{arydshln}
\usepackage{mathtools}
\usepackage{mathrsfs}
\usepackage{algorithm}
\usepackage{algorithmic}
\usepackage{graphicx}
\usepackage{epsfig,graphics}
\usepackage{epstopdf}
\usepackage{subfigure}
\usepackage{enumitem}
\usepackage{pifont}
\usepackage{multirow,ragged2e}
\usepackage{bm}
\usepackage{fixltx2e}
\usepackage[english]{babel}
\usepackage{color}
\usepackage[table]{xcolor}
\usepackage{cuted}
\usepackage{stfloats}
\usepackage{orcidlink}

\usepackage[percent]{overpic}
\usepackage{pict2e}

\usepackage{siunitx}

\usepackage{url}

\usepackage{tikz}

\usepackage{filecontents}
\theoremstyle{theorem}

\theoremstyle{lemma}

\theoremstyle{definition}

\theoremstyle{assumption}

\theoremstyle{problem}

\theoremstyle{example}

\theoremstyle{proposition}

\theoremstyle{corollary}

\theoremstyle{property}

\theoremstyle{remark}

\newcommand{\fig}[1]{Fig.~\ref{#1}}

\newcommand{\tab}[1]{Table~\ref{#1}}
\newcommand{\sect}[1]{Section~\ref{#1}}

\newcommand{\eq}[1]{(\ref{#1})}
\newcommand{\eqs}[2]{Equations~(\ref{#1})-(\ref{#2})}

\newcommand{\fbm}[1]{\mathbf{#1}}
\newcommand{\tbm}[1]{\fbm{#1}^\mathsf{T}}
\newcommand{\tfbm}[1]{\bm{#1}^\mathsf{T}}
\newcommand{\ibm}[1]{\fbm{#1}^{-1}}

\newcommand{\bbm}[1]{\overline{\fbm{#1}}}
\newcommand{\tbbm}[1]{\overline{\fbm{#1}}{}^\mathsf{T}}
\newcommand{\ibbm}[1]{\overline{\fbm{#1}}{}^\mathsf{-1}}

\newcommand{\wtilbm}[1]{\widetilde{\fbm{#1}}}
\newcommand{\twtilbm}[1]{\widetilde{\fbm{#1}}{}^\mathsf{T}}

\newcommand{\dottbm}[1]{\dot{\fbm{#1}}{}^\mathsf{T}}
\newcommand{\dotbm}[1]{\dot{\fbm{#1}}}

\newcommand{\dotwtilbm}[1]{\dot{\widetilde{\fbm{#1}}}}

\newcommand{\dottwtilbm}[1]{\dot{\widetilde{\fbm{#1}}}{}^\mathsf{T}}

\newcommand{\sk}[0]{\text{sk}}
\newcommand{\tr}[0]{\text{tr}}

\usepackage{caption}
\begin{document}

\title{Image-Based Visual Servoing for Enhanced Cooperation of Dual-Arm Manipulation}

\author{Zizhe Zhang\,\orcidlink{0009-0009-0662-0327},~\IEEEmembership{Graduate Student Member,~IEEE}, Yuan Yang\,\orcidlink{0000-0002-2317-9002},~\IEEEmembership{Member,~IEEE}, Wenqiang Zuo, Guangming Song\,\orcidlink{0000-0002-6362-4043},~\IEEEmembership{Senior Member,~IEEE}, Aiguo Song\,\orcidlink{0000-0002-1982-6780},~\IEEEmembership{Senior Member,~IEEE}, Yang Shi\,\orcidlink{0000-0003-1337-5322},~\IEEEmembership{Fellow,~IEEE}
        \thanks{Zizhe Zhang is with the Department of Electrical and System Engineering, University of Pennsylvania, Philadelphia, PA 19104, United States (e-mail: zizhez@seas.upenn.edu).}
        \thanks{Yuan Yang, Guangming Song and Aiguo Song are with the School of Instrument Science and Engineering, Southeast University, Nanjing, Jiangsu 210096, China (e-mail: yuan\underline{ }evan\underline{ }yang@seu.edu.cn; mikesong@seu.edu.cn; a.g.song@seu.edu.cn).}
        \thanks{Wenqiang Zuo is with the School of Materials Science and Engineering, Southeast University, Nanjing, Jiangsu 211189, China (e-mail: wenqiangzuo@seu.edu.cn).}
        \thanks{Yang Shi is with the Department of Mechanical Engineering, University of Victoria, Victoria, BC V8W 2Y2, Canada (e-mail: yshi@uvic.ca).}
        \thanks{This article has supplementary downloadable material available at \href{https://doi.org/10.1109/LRA.2025.3543137}{https://doi.org/10.1109/LRA.2025.3543137}, provided by the authors.}
        \thanks{Digital Object Identifier 10.1109/LRA.2025.3543137}
        }

\maketitle

\begin{abstract}
The cooperation of a pair of robot manipulators is required to manipulate a target object without any fixtures. The conventional control methods coordinate the end-effector pose of each manipulator with that of the other using their kinematics and joint coordinate measurements. Yet, the manipulators' inaccurate kinematics and joint coordinate measurements can cause significant pose synchronization errors in practice. This paper thus proposes an image-based visual servoing approach for enhancing the cooperation of a dual-arm manipulation system. On top of the classical control, the visual servoing controller lets each manipulator use its carried camera to measure the image features of the other's marker and adapt its end-effector pose with the counterpart on the move. Because visual measurements are robust to kinematic errors, the proposed control can reduce the end-effector pose synchronization errors and the fluctuations of the interaction forces of the pair of manipulators on the move. Theoretical analyses have rigorously proven the stability of the closed-loop system. Comparative experiments on real robots have substantiated the effectiveness of the proposed control.
\end{abstract}


\section{Introduction}

{Dual-arm manipulation is a challenging research topic in the robotics society~\cite{Billard2019Science,Abbas2023IJIRA}.} Given that typical humanoid robots have bimanual capabilities, the study of dual-arm manipulation control is gaining more and more attention recently~\cite{Rakita2019Science}. A typical application of dual-arm manipulation is to grasp and transport a rigid target object without extra fixtures~\cite{Tamin2022RAL}. In this context, the two arms need to keep their end effectors in frictional contact with the manipulated object throughout the task execution~\cite{Caccavale2016Springer}. Thus, a fundamental control objective for the pair of two manipulators is to tightly coordinate their end-effector poses on the move~\cite{Siciliano1996JMSE}.

\begin{figure}[!ht]
\centering
\begin{overpic}[width=\columnwidth]{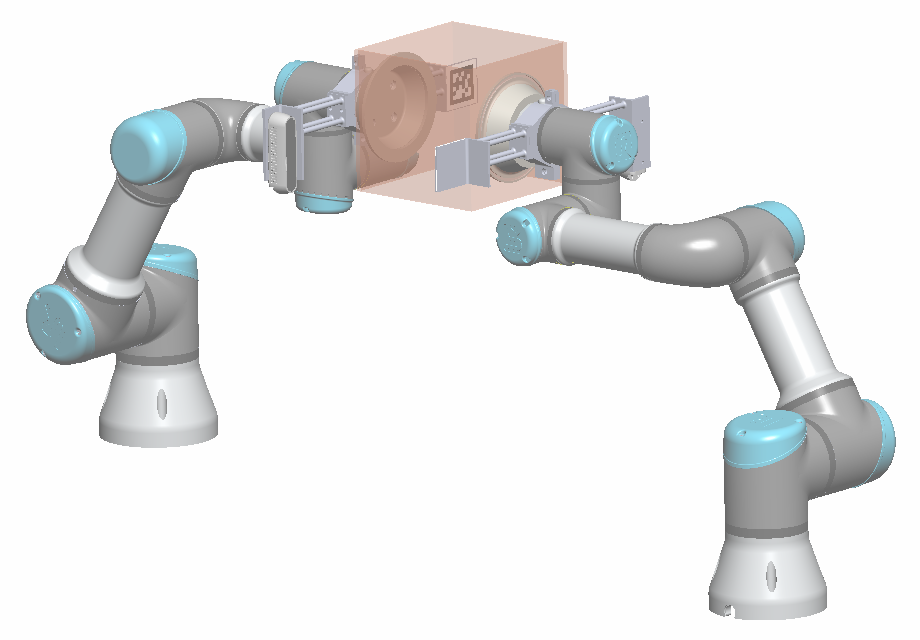}
\linethickness{1pt}
\put(11,17){\footnotesize\textcolor{black}{Left Arm}}
\put(60,5){\footnotesize\textcolor{black}{Right Arm}}
\put(42,11){\footnotesize\textcolor{black}{$\mathcal{F}_{w}$}}
\put(45,15){\color{red}\vector(1,0.4){7}}
\put(45,15){\color{green}\vector(-1,0.4){7}}
\put(45,15){\color{blue}\vector(0,1){8}}
\put(12,24){\footnotesize\textcolor{black}{$\mathcal{F}_{b\ast}$}}
\put(17,26){\color{red}\vector(1,-0.1){6}}
\put(17,26){\color{green}\vector(1,0.2){6}}
\put(17,26){\color{blue}\vector(0,1){6}}
\put(36,58){\footnotesize\textcolor{black}{$\mathcal{F}_{e\ast}$}}
\put(42,58){\color{red}\vector(0,-1){6}}
\put(42,58){\color{green}\vector(1,0.2){6}}
\put(42,58){\color{blue}\vector(1,-0.1){6}}
\put(26,54){\footnotesize\textcolor{black}{$\mathcal{F}_{c\ast}$}}
\put(31,53){\color{red}\vector(0,-1){6}}
\put(31,53){\color{green}\vector(1,0.2){6}}
\put(31,53){\color{blue}\vector(1,-0.1){6}}
\put(52,52){\footnotesize\textcolor{black}{$\mathcal{F}_{o}$}}
\put(50,56){\color{red}\vector(0,-1){6}}
\put(50,56){\color{green}\vector(1,0.2){6}}
\put(50,56){\color{blue}\vector(1,-0.1){6}}
\put(23,18){\tikz \draw[-latex,black] (0,0) arc [start angle=15,end angle=105,x radius=1.5cm,y radius=1.25cm];}
\put(30,23){\footnotesize\textcolor{black}{$\fbm{T}_{b\ast}$}}
\put(45.5,18){\tikz \draw[-latex,black] (0,0) arc [start angle=-20,end angle=20,x radius=5.0cm,y radius=4.75cm];}
\put(43,36){\footnotesize\textcolor{black}{$\fbm{T}_{\ast}$}}
\put(47,18){\tikz \draw[-latex,black] (0,0) arc [start angle=-30,end angle=5,x radius=5.0cm,y radius=5.0cm];}
\put(55,30){\footnotesize\textcolor{black}{$\fbm{T}_{o}$}}
\put(22,30){\tikz \draw[-latex,black] (0,0) arc [start angle=-60,end angle=0,x radius=3.85cm,y radius=2.5cm];}
\put(29,39){\footnotesize\textcolor{black}{$\fbm{T}^{b}_{\ast}$}}
\put(42,56){\tikz \draw[-latex,black] (0,0) arc [start angle=145,end angle=0,x radius=0.4cm,y radius=0.45cm];}
\put(44,62.5){\footnotesize\textcolor{black}{$\fbm{T}^{\ast}_{o}$}}
\end{overpic}
\caption{Two $6$-DOF serial manipulators serve as the left and the right arms in a dual-arm manipulation system. The target object that they manipulate is a cubic box. $\mathcal{F}_w$ and $\mathcal{F}_o$ denote the world frame and the object frame, respectively. $\mathcal{F}_{b\ast}$, $\mathcal{F}_{e\ast}$ and $\mathcal{F}_{c\ast}$ are the base, end-effector and camera frames of the left~$\ast=l$ and the right~$\ast=r$ arms. $\fbm{T}_{o}$, $\fbm{T}_{b\ast}$, $\fbm{T}_{\ast}$, $\fbm{T}^{b}_{\ast}$ and $\fbm{T}^{\ast}_{o}$ indicate the transformations between the frames.}
\label{fig:frame}
\end{figure}

The existing research on dual-arm manipulation control has synchronized the end-effector positions of both manipulators with event-triggered communications~\cite{Hirche2020TRO} and regulated their interaction forces with the object in the contact direction~\cite{Haddadin2022RAL}. Accounting for the internal forces~\cite{Walker1991IJRR}, constrained quadratic programs have been formulated to minimize the contact forces while avoiding obstacles in the surrounding environment~\cite{StevenLiu2021Mechatronics} and the separation or sliding of the frictional contact~\cite{Dehio2021IJRR}. Under nonlinear model predictive control, a torque-controlled dual-arm manipulation system has recently been tested for a mirror grinding task~\cite{HanDing2023TIE}. However, torque-controlled manipulators with exactly known dynamic models are unavailable in many applications. Cooperation approaches for velocity- and position-controlled robot arms have thus been advocated. For velocity-controlled manipulators, the joint limit avoidance and Cartesian-space path following have been pursued with lower priority than their pose synchronization task~\cite{Kyrki2018RAS}. A sparse kinematic control has minimized the number of the two arms' actuated joints using a hierarchical quadratic program~\cite{Sonny2018IROS}. For position-controlled manipulators, a linear elastic model for the contact of the follower arm with the object has been assumed to minimize the internal forces~\cite{Costanzo2022RAL}. A communication-free coordination control has adopted the manipulators' interaction forces to transport deformable objects~\cite{Gombo2024TRO}.

A position-based visual servoing~(PBVS) approach has been deployed on a humanoid robot to control both arms/hands to grasp and manipulate its observed target objects in a kitchen environment~\cite{Nikolaus2009ICHR}. An image-based visual servoing~(IBVS) method has been applied to a dual-arm robot with both fixed and mobile cameras to align the poses of both arms when performing a plug-socket insertion task~\cite{YiliFu2017Robotica}. Yet, the control design and stability analysis have not considered the closed kinematic chain. With IBVS, two simulated aerial manipulators have rigidly grasped an assembly bar and transported it to the target pose~\cite{Siciliano2015IROS}. However, the control has not yet been verified on any real-robot platform.

\subsection{Contributions}
This paper develops an IBVS control algorithm for enhancing dual-arm cooperation. Its contributions to the coordination control and the visual servoing control are as follows.

The classical cooperative manipulation controls synchronize the end-effector poses of both arms~\cite{Spong2012TAC}. However, previous studies~\cite{Yuan2022TRO,Yuan2025TRO} have discovered that the kinematic errors of manipulators can induce significant pose biases between them and lead to possible task failures and hardware damage. Recursive least-squares estimators have estimated unknown kinematic parameters of space manipulators while suffering from joint measurement noises~\cite{Aghili2013TMECH}. Because IBVS is robust to kinematic errors, the paper proposes an IBVS algorithm to reduce the pose errors and the accompanying interaction force fluctuations during dual-arm manipulation.

The basic~\cite{Chaumette2006RAM} and advanced~\cite{Chaumette2007RAM} visual servoing algorithms have been widely employed for a single manipulator to reach a stationary target or track a moving target. The interaction matrices depend only on the image features and their depths. However, in the context of dual-arm manipulation, the cameras and markers are mounted on two arms separately. As a result, the kinematic model of the system becomes entirely different, and the associated interaction matrix significantly differs from the classical one. The paper figures out the dependence of the customized interaction matrix on the poses of both arms explicitly. Moreover, it synthesizes the IBVS algorithm with pose synchronization control while guaranteeing the closed-loop system stability rigorously.

In the remainder of the paper, \sect{sec: modelling} builds the kinematic models of the manipulators and the vision system. \sect{sec: control design and stability analysis} designs the control algorithm and analyses the system stability. \sect{sec: discussion} discusses the state and input constraints of manipulators and some visual servoing issues. \sect{sec: experimental results} validates the control via comparative experiments. Finally, \sect{sec: conclusion} concludes the design and future studies.
\color{black}

\section{Modelling}\label{sec: modelling}

This section introduces the closed kinematic chain formed by the pair of manipulators and the pinhole perspective projection model of their wrist-mounted cameras.

\subsection{The Closed Kinematic Chain}

The dual-arm manipulation system, as shown in \fig{fig:frame}, contains a pair of $6$-degrees-of-freedom ($6$-DOF) serial manipulators with only revolute joints. Let the world frame be $\mathcal{F}_{w}$, the base and the end-effector frames of the left~$\ast=l$ and the right~$\ast=r$ arms be $\mathcal{F}_{b\ast}$ and $\mathcal{F}_{e\ast}$, respectively. Let $\fbm{q}_{\ast}$ be the joint coordinates and
\begin{align*}
\fbm{T}_{\ast}=\begin{bmatrix}\fbm{R}_{\ast} &\fbm{x}_{\ast}\\ \fbm{0} &1\end{bmatrix}
\end{align*}
be the end-effector poses of the two arms~$\ast=l,r$, where $\fbm{R}_{\ast}$ and $\fbm{x}_{\ast}$ are the rotation matrix representations of orientation and the positions respectively. The Jacobian matrices~$\fbm{J}_{\ast}$ of the two manipulators~$\ast=l,r$ then link their joint velocity inputs~$\dotbm{q}_{\ast}$ to their end-effector velocities~$\fbm{v}_{\ast}$ by
\begin{equation}\label{equ1}
\fbm{v}_{\ast}=\begin{bmatrix}\bm{\upsilon}_{\ast}\\ \bm{\omega}_{\ast}\end{bmatrix}=\fbm{J}_{\ast}\dotbm{q}_{\ast}\textrm{,}
\end{equation}
where the linear~$\bm{\upsilon}_{\ast}$ and angular~$\bm{\omega}_{\ast}$ velocities adapt their end-effector poses~$\fbm{T}_{\ast}$ by
\begin{subequations}\label{equ2}
\begin{align}
\dotbm{x}_{\ast}=&\bm{\upsilon}_{\ast}\textrm{,}\label{equ2a}\\
\dotbm{R}_{\ast}=&\bm{\omega}^{\times}_{\ast}\fbm{R}_{\ast}\textrm{.}\label{equ2b}
\end{align}
\end{subequations}

To manipulate the target object without fixtures, the two manipulators need to keep their end effectors rigidly attached to the object throughout the task execution. Let $\mathcal{F}_{o}$ be the object frame, and let $\fbm{T}_{o}$ be the object pose. Then, the rigid attachment of the arms' end effectors to the object requires
\begin{equation}\label{equ3}
\fbm{T}_{l}\fbm{T}^{l}_{o}=\fbm{T}_{o}=\fbm{T}_{r}\fbm{T}^{r}_{o}\textrm{,}
\end{equation}
where the poses of the object relative to the end effectors of the left~$\fbm{T}^{l}_{o}$ and the right~$\fbm{T}^{r}_{o}$ arms remain invariant. Simple algebraic operations derive that the transformations
\begin{subequations}\label{equ4}
\begin{align}
\fbm{T}^{l}_{r}=&\ibm{T}_{l}\fbm{T}_{r}=\fbm{T}^{l}_{o}\fbm{T}^{r}_{o}{}^{-1}\textrm{,}\label{equ4a}\\
\fbm{T}^{r}_{l}=&\ibm{T}_{r}\fbm{T}_{l}=\fbm{T}^{r}_{o}\fbm{T}^{l}_{o}{}^{-1}\label{equ4b}
\end{align}
\end{subequations}
remain invariant. Namely, the two arms' end effectors and the object shall stay stationary with respect to each other. For the left arm~$\ast=l$, the right arm~$\ast=r$, and the object~$\ast=o$, let
\begin{align*}
\bbm{T}_{\ast}=\begin{bmatrix}\bbm{R}_{\ast} &\bbm{x}_{\ast}\\ \fbm{0} &1\end{bmatrix}\quad \text{and}\quad \fbm{T}_{\ast d}=\begin{bmatrix}\fbm{R}_{\ast d} &\fbm{x}_{\ast d}\\ \fbm{0} &1\end{bmatrix}
\end{align*}
denote their initial and desired poses, respectively. The pose synchronization control objective of the dual-arm manipulation system can thus be formulated as
\begin{subequations}\label{equ5}
\begin{align}
\ibm{T}_{od}\fbm{T}_{o}=&\fbm{I}\textrm{,}\label{equ5a}\\
\ibbm{T}_{l}\fbm{T}_{l}=\ibbm{T}_{r}\fbm{T}_{r}=&\ibbm{T}_{o}\fbm{T}_{o}\textrm{.}\label{equ5b}
\end{align}
\end{subequations}

In \eq{equ5a}, $\ibm{T}_{od}\fbm{T}_{o}$ is the displacement of the object from its desired pose~$\fbm{T}_{od}$. It thus enforces the object to move towards its desired pose. In \eq{equ5b}, $\ibbm{T}_{\ast}\fbm{T}_{\ast}$ denote the displacements of the left arm~$\ast=l$, the right arm~$\ast=r$, and the object~$\ast=o$ from their initial poses~$\bbm{T}_{\ast}$, respectively. As the two manipulators reliably hold the object at the initial poses~$\bbm{T}_{\ast}$, the objective~\eqref{equ5b} then secures the manipulation by enforcing them to keep their relative poses invariant.

\color{black}

\begin{figure}[!ht]
\centering
\begin{overpic}[width=.95\columnwidth]{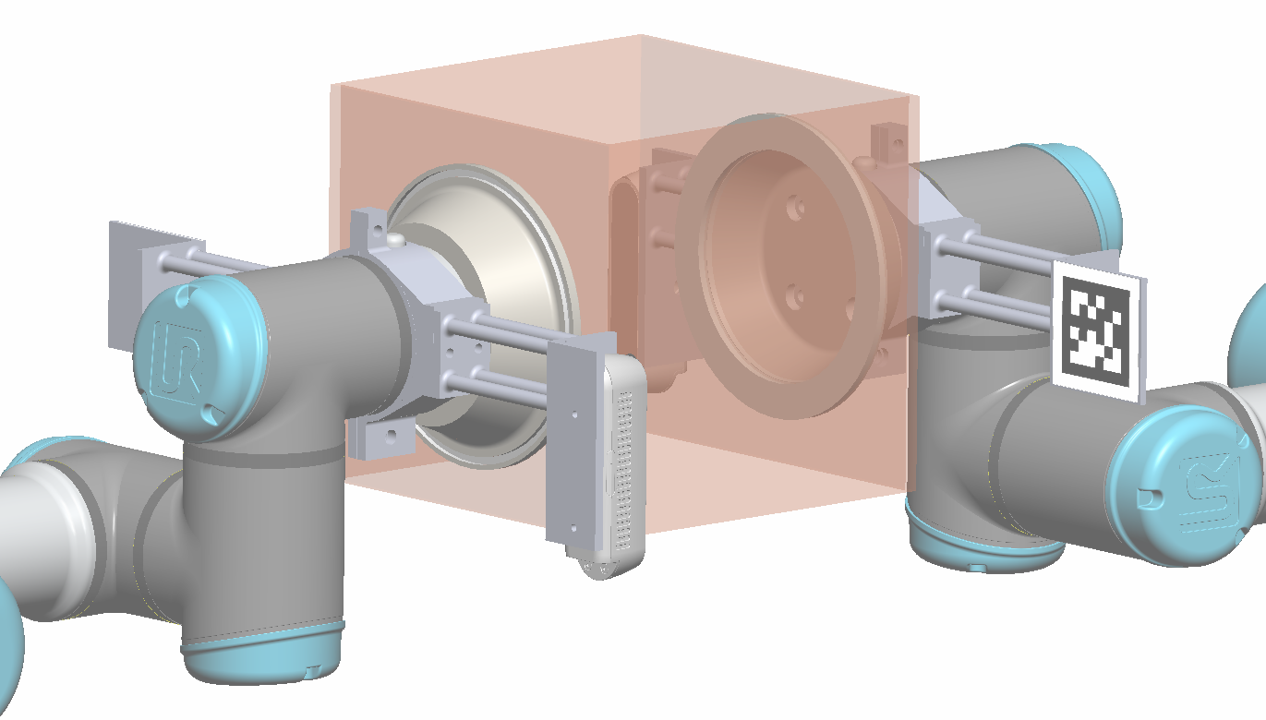}
\linethickness{1pt}
\put(3,0){\footnotesize\textcolor{black}{Left Arm}}
\put(85,9){\footnotesize\textcolor{black}{Right Arm}}
\put(39,0){\footnotesize\textcolor{black}{$\mathcal{F}_{w}$}}
\put(50,0){\color{red}\vector(-1,0.4){7}}
\put(50,0){\color{green}\vector(-1,-0.4){7}}
\put(50,0){\color{blue}\vector(0,1){8}}
\put(43,17){\footnotesize\textcolor{black}{$\mathcal{F}_{cl}$}}
\put(49,20){\color{red}\vector(0,-1){8}}
\put(49,20){\color{green}\vector(-1,0.3){7}}
\put(49,20){\color{blue}\vector(1,0.4){7}}
\put(37,36){\footnotesize\textcolor{black}{$\mathcal{F}_{el}$}}
\put(40,33){\color{red}\vector(0,-1){8}}
\put(40,33){\color{green}\vector(-1,0.3){7}}
\put(40,33){\color{blue}\vector(1,0.4){7}}
\put(60,40){\footnotesize\textcolor{black}{$\mathcal{F}_{er}$}}
\put(64,38){\color{red}\vector(0,-1){8}}
\put(64,38){\color{green}\vector(1,-0.3){7}}
\put(64,38){\color{blue}\vector(-1,-0.4){7}}
\put(90,23.0){\footnotesize\textcolor{black}{$1$}}
\put(89,26.5){\color{brown}\circle*{1.5}}
\put(90,35.5){\footnotesize\textcolor{black}{$2$}}
\put(89,34.0){\color{brown}\circle*{1.5}}
\put(84,36.5){\footnotesize\textcolor{black}{$3$}}
\put(84,35.0){\color{brown}\circle*{1.5}}
\put(84,24.0){\footnotesize\textcolor{black}{$4$}}
\put(84,27.5){\color{brown}\circle*{1.5}}
\linethickness{0.5pt}
\put(49,20){\color{black}\vector(1,0.21){35}}
\put(67,21){\footnotesize\textcolor{black}{$\fbm{m}_{l4}$}}
\put(64,38){\color{black}\vector(1,-0.5){20}}
\put(67,32){\footnotesize\textcolor{black}{$\fbm{p}_{r4}$}}
\put(40,21){\tikz \draw[-latex,black] (0,0) arc [start angle=75,end angle=15,x radius=1.05cm,y radius=1.45cm];}
\put(47,29){\footnotesize\textcolor{black}{$\fbm{T}^{l}_{c}$}}
\put(33,0){\tikz \draw[-latex,black] (0,0) arc [start angle=-120,end angle=-205,x radius=2.2cm,y radius=2.1cm];}
\put(33,10){\footnotesize\textcolor{black}{$\fbm{T}_{l}$}}
\put(50,0){\tikz \draw[-latex,black] (0,0) arc [start angle=-75,end angle=-10,x radius=1.5cm,y radius=3.3cm];}
\put(60,10){\footnotesize\textcolor{black}{$\fbm{T}_{r}$}}
\end{overpic}
\caption{The positions of the corner points~$i=1,2,3,4$ expressed of the right arm's marker expressed in the end-effector frame~$\mathcal{F}_{er}$ and in the left camera frame~$\mathcal{F}_{cl}$ are $\fbm{p}_{ri}$ and $\fbm{m}_{li}$, respectively. $\fbm{T}_{\ast}$ and $\fbm{T}^{l}_{c}$ denote the transformations between the world frame~$\mathcal{F}_{w}$, the end-effector frames~$\mathcal{F}_{e\ast}$ and the camera frame~$\mathcal{F}_{cl}$ with $\ast=l,r$.}
\label{fig:marker}
\end{figure}

\subsection{The Pinhole Projection Model}
The eye-in-hand camera mounted on the wrist of each manipulator, as shown in \fig{fig:marker}, measures the image features of the fiducial marker fixed on the wrist of the other manipulator. In particular, the image features adopted in the control design are the four corner points of each AprilTag marker. Let the pose of the camera frame~$\mathcal{F}_{c\ast}$ relative to the end-effector frame~$\mathcal{F}_{e\ast}$ of the left~$\ast=l$ or the right~$\ast=r$ arm be
\begin{align*}
\fbm{T}^{\ast}_{c}=\begin{bmatrix}\fbm{R}^{\ast}_{c} &\fbm{x}^{\ast}_{c}\\ \fbm{0} &1\end{bmatrix}\textrm{.}
\end{align*}
Let the position of the $i$-th corner point of the marker on the right~$\ast=r$ or the left~$\ast=l$ arm be expressed in the end-effector frame~$\mathcal{F}_{e\ast}$ by $\fbm{p}_{\ast i}$ with $i=1,2,3,4$.

{Let the superscripts~${}^x$, ${}^y$  and ${}^z$ index the components of position vectors in the associated axis.} By the kinematic chain, the position~$\fbm{m}_{li}=(m^{x}_{li},m^{y}_{li},m^{z}_{li})^\mathsf{T}$ of the $i$-th corner point of the right arm's marker expressed in the left arm's camera frame~$\mathcal{F}_{cl}$ meets
\begin{align*}
\fbm{T}_{l}\fbm{T}^{l}_{c}\begin{bmatrix}\fbm{m}_{li}\\ 1\end{bmatrix}=\fbm{T}_{r}\begin{bmatrix}\fbm{p}_{ri}\\ 1\end{bmatrix}\textrm{.}
\end{align*}
The position~$\fbm{m}_{ri}=(m^{x}_{ri},m^{y}_{ri},m^{z}_{ri})^\mathsf{T}$ of the $i$-th corner point of the left arm's marker expressed in the right arm's camera frame~$\mathcal{F}_{r}$ meets
\begin{align*}
\fbm{T}_{r}\fbm{T}^{r}_{c}\begin{bmatrix}\fbm{m}_{ri}\\ 1\end{bmatrix}=\fbm{T}_{l}\begin{bmatrix}\fbm{p}_{li}\\ 1\end{bmatrix}\textrm{.}
\end{align*}
One can thus derive that
\begin{subequations}\label{equ6}
\begin{align}
\fbm{m}_{li}=&\left(\fbm{R}_{l}\fbm{R}^{l}_{c}\right)^\mathsf{T}\left(\fbm{R}_{r}\fbm{p}_{ri}+\fbm{x}_{r}-\fbm{R}_{l}\fbm{x}^{l}_{c}-\fbm{x}_{l}\right)\textrm{,}\label{equ6a}\\
\fbm{m}_{ri}=&\left(\fbm{R}_{r}\fbm{R}^{r}_{c}\right)^\mathsf{T}\left(\fbm{R}_{l}\fbm{p}_{li}+\fbm{x}_{l}-\fbm{R}_{r}\fbm{x}^{r}_{c}-\fbm{x}_{r}\right)\textrm{,}\label{equ6b}
\end{align}
\end{subequations}
for the corner points~$i=1,2,3,4$ of markers.

Let $\bm{\zeta}_{\ast i}$ normalize $\fbm{m}_{\ast i}$ with $\ast=l,r$ by their $z$-coordinates
\begin{align*}
\bm{\zeta}_{\ast i}=\begin{bmatrix}\frac{m^{x}_{\ast i}}{m^{z}_{\ast i}} &\frac{m^{y}_{\ast i}}{m^{z}_{\ast i}}\end{bmatrix}^\mathsf{T}\textrm{.}
\end{align*}
Let the intrinsic parameters matrices of the left~$\ast=l$ and the right~$\ast=r$ cameras be $\fbm{K}_{\ast}$. The cameras then capture their image features~$\fbm{s}_{\ast i}$ with $i=1,2,3,4$ by
\begin{equation}\label{equ7}
\fbm{s}_{\ast i}=\fbm{K}_{\ast}\bm{\zeta}_{\ast i}\textrm{.}
\end{equation}
Taking the time derivative of \eq{equ7} yields
\begin{equation}\label{equ8}
\dotbm{s}_{\ast i}=\frac{1}{m^{z}_{\ast i}}\begin{bmatrix}\fbm{K}_{\ast} &-\fbm{s}_{\ast i}\end{bmatrix}\dotbm{m}_{\ast i}\textrm{.}
\end{equation}
Substituting \eq{equ6} into \eq{equ8} gives rise to
\begin{equation}
\begin{bmatrix}\dotbm{s}_{li}\\ \dotbm{s}_{ri}\end{bmatrix}=\begin{bmatrix}\fbm{L}^{l}_{li} &\fbm{L}^{r}_{li}\\ \fbm{L}^{l}_{ri} &\fbm{L}^{r}_{ri}\end{bmatrix}\begin{bmatrix}\fbm{v}_{l}\\ \fbm{v}_{r}\end{bmatrix}\textrm{,}
\end{equation}
where the interaction matrices~$\fbm{L}^{l}_{li}$, $\fbm{L}^{r}_{li}$, $\fbm{L}^{l}_{ri}$ and $\fbm{L}^{r}_{ri}$ with $i=1,2,3,4$ are expressed as
\begin{align*}
\fbm{L}^{l}_{li}=&\frac{1}{{m^{z}_{li}}}\begin{bmatrix}\fbm{K}_{l}\ -\fbm{s}_{li}\end{bmatrix}\left(\fbm{R}_{l}\fbm{R}^{l}_{c}\right)^\mathsf{T}\begin{bmatrix}-\fbm{I} &\left(\fbm{R}_{r}\fbm{p}_{ri}+\fbm{x}_{r}-\fbm{x}_{l}\right)^{\times}\end{bmatrix}\textrm{,}\\
\fbm{L}^{r}_{li}=&\frac{1}{{m^{z}_{li}}}\begin{bmatrix}\fbm{K}_{l}\ -\fbm{s}_{li}\end{bmatrix}\left(\fbm{R}_{l}\fbm{R}^{l}_{c}\right)^\mathsf{T}\begin{bmatrix}\fbm{I} &-\left(\fbm{R}_{r}\fbm{p}_{ri}\right)^{\times}\end{bmatrix}\textrm{,}\\
\fbm{L}^{l}_{ri}=&\frac{1}{{m^{z}_{ri}}}\begin{bmatrix}\fbm{K}_{r}\ -\fbm{s}_{ri}\end{bmatrix}\left(\fbm{R}_{r}\fbm{R}^{r}_{c}\right)^\mathsf{T}\begin{bmatrix}\fbm{I} &-\left(\fbm{R}_{l}\fbm{p}_{li}\right)^{\times}\end{bmatrix}\textrm{,}\\
\fbm{L}^{r}_{ri}=&\frac{1}{{m^{z}_{ri}}}\begin{bmatrix}\fbm{K}_{r}\ -\fbm{s}_{ri}\end{bmatrix}\left(\fbm{R}_{r}\fbm{R}^{r}_{c}\right)^\mathsf{T}\begin{bmatrix}-\fbm{I} &\left(\fbm{R}_{l}\fbm{p}_{li}+\fbm{x}_{l}-\fbm{x}_{r}\right)^{\times}\end{bmatrix}\textrm{.}
\end{align*} 
Let $\bbm{s}_{li}$ and $\bbm{s}_{ri}$ be the initial values of $\fbm{s}_{li}$ and $\fbm{s}_{ri}$, respectively. The enhanced cooperation control objective of the dual-arm system can thus be formulated as
\begin{equation}\label{equ10}
\begin{bmatrix}\fbm{s}_{li}\\ \fbm{s}_{ri}\end{bmatrix}=\begin{bmatrix}\bbm{s}_{li}\\ \bbm{s}_{ri}\end{bmatrix}
\end{equation}
for the corner points~$i=1,2,3,4$ of markers.

\section{Control Design and Stability Analysis}\label{sec: control design and stability analysis}

This section designs the control inputs to the manipulators and analyses the stability of the closed-loop system. 

\subsection{Control Design}
\eq{equ5b} implies that
\begin{align*}
\ibbm{T}_{l}\fbm{T}_{ld}=\ibbm{T}_{r}\fbm{T}_{rd}=\ibbm{T}_{o}\fbm{T}_{od}\textrm{.}
\end{align*}
Given that $\fbm{T}_{od}$ is pre-planned, one can then distribute desired poses to the left and the right arms by
\begin{align*}
\fbm{T}_{ld}=\bbm{T}_{l}\ibbm{T}_{o}\fbm{T}_{od}\quad \text{and}\quad \fbm{T}_{rd}=\bbm{T}_{r}\ibbm{T}_{o}\fbm{T}_{od}\textrm{.}
\end{align*}
As a result, \eq{equ5a} is equivalent to
\begin{align*}
\ibm{T}_{\ast d}\fbm{T}_{\ast}=\begin{bmatrix}\tbm{R}_{\ast d}\fbm{R}_{\ast} &\tbm{R}_{\ast d}(\fbm{x}_{\ast}-\fbm{x}_{\ast d})\\ \fbm{0} &1\end{bmatrix}=\fbm{I}
\end{align*}
for $\ast=l,r$ and can thus be reformulated as
\begin{equation}\label{equ11}
\mathop{\text{minimize}}\limits_{\fbm{q}_{l},\fbm{q}_{r}}\quad V_{d}\textrm{,}
\end{equation}
where the objective function is defined by
\begin{align*}
V_{d}=\frac{1}{2}\sum_{\ast=l,r}\Big[\tr\left(\fbm{I}-\tbm{R}_{\ast d}\fbm{R}_{\ast}\right)+(\fbm{x}_{\ast}-\fbm{x}_{\ast d})^\mathsf{T}(\fbm{x}_{\ast}-\fbm{x}_{\ast d})\Big]
\end{align*}
with $\tr(\fbm{M})$ being the trace of any square matrix~$\fbm{M}$.
\color{black}

\eq{equ5b} implies also that
\begin{align*}
\bbm{T}{}^{-1}_{l}\fbm{T}_{l}=\begin{bmatrix}\wtilbm{R}_{l} &\tbbm{R}_{l}\wtilbm{x}_{l}\\ \fbm{0} &1\end{bmatrix}=\begin{bmatrix}\wtilbm{R}_{r} &\tbbm{R}_{r}\wtilbm{x}_{r}\\ \fbm{0} &1\end{bmatrix}=\bbm{T}{}^{-1}_{r}\fbm{T}_{r}\textrm{,}
\end{align*}
where $\wtilbm{R}_{\ast}=\tbbm{R}_{\ast}\fbm{R}_{\ast}$ and $\wtilbm{x}_{\ast}=\fbm{x}_{\ast}-\bbm{x}_{\ast}$ for $\ast=l,r$. Namely,
\begin{align*}
\fbm{I}=\wtilbm{R}_{l}\twtilbm{R}_{r}\quad \text{and}\quad \fbm{0}=\tbbm{R}_{l}\wtilbm{x}_{l}-\tbbm{R}_{r}\wtilbm{x}_{r}\textrm{.}
\end{align*}
It can thus be reformulated as follows
\begin{equation}\label{equ12}
\mathop{\text{minimize}}\limits_{\fbm{q}_{l},\fbm{q}_{r}}\quad V_t\textrm{,}
\end{equation}
where the objective function is defined by
\begin{align*}
V_{t}=\frac{1}{2}\tr\big(\fbm{I}-\wtilbm{R}_{l}\twtilbm{R}_{r}\big)+\frac{1}{2}\left(\tbbm{R}_{l}\wtilbm{x}_{l}-\tbbm{R}_{r}\wtilbm{x}_{r}\right)^\mathsf{T}\left(\tbbm{R}_{l}\wtilbm{x}_{l}-\tbbm{R}_{r}\wtilbm{x}_{r}\right)\textrm{.}
\end{align*}
{The enhanced cooperation control objective~\eqref{equ10} can be reformulated as the following quadratic program}
\begin{equation}\label{equ13}
\mathop{\text{minimize}}\limits_{\fbm{q}_{l},\fbm{q}_{r}}\quad V_{s}\textrm{,}
\end{equation}
where the objective function is defined by
\begin{align*}
V_{s}=\frac{1}{2}\sum^{4}_{i=1}\left[(\fbm{s}_{li}-\bbm{s}_{li})^\mathsf{T}(\fbm{s}_{li}-\bbm{s}_{li})+(\fbm{s}_{ri}-\bbm{s}_{ri})^\mathsf{T}(\fbm{s}_{ri}-\bbm{s}_{ri})\right]\textrm{.}
\end{align*}
Therefore, IBVS for enhancing the cooperation of dual-arm manipulation is formulated as solving the optimization problems~\eqref{equ11}-\eqref{equ13} simultaneously by
\begin{equation}\label{equ14}
\mathop{\text{minimize}}\limits_{\fbm{q}_{l},\fbm{q}_{r}}\quad V={w_{d}V_{d}+}w_{t}V_{t}+w_{s}V_{s}\textrm{,}
\end{equation}
where the objective function~$V$ sums $V_{d}$, $V_{t}$ and $V_{s}$ by positive scalar weights~$w_{d}$, $w_{t}$ and $w_{s}$, respectively.

Following the steepest descent algorithm, the joint velocity control inputs to the left and the right arms are designed by \eqref{equ15},
\begin{figure*}[b]
\hrulefill
\begin{subequations}\label{equ15}
\begin{align}
\dotbm{q}_{l}=&{-w_{d}\tbm{J}_{l}\begin{bmatrix}\fbm{x}_{l}-\fbm{x}_{ld}\\ \sk\left(\fbm{R}_{l}\tbm{R}_{ld}\right)^{\vee}\end{bmatrix}}-w_{t}\tbm{J}_{l}\begin{bmatrix}\wtilbm{x}_{l}-\bbm{R}_{l}\tbbm{R}_{r}\wtilbm{x}_{r}\\ \sk\left(\fbm{R}_{l}\twtilbm{R}_{r}\tbbm{R}_{l}\right)^{\vee}\end{bmatrix}-w_{s}\tbm{J}_{l}\sum^{4}_{i=1}\Big[\fbm{L}^{l}_{li}{}^\mathsf{T}\left(\fbm{s}_{li}-\bbm{s}_{li}\right)+\fbm{L}^{l}_{ri}{}^\mathsf{T}\left(\fbm{s}_{ri}-\bbm{s}_{ri}\right)\Big]\\
\dotbm{q}_{r}=&{-w_{d}\tbm{J}_{r}\begin{bmatrix}\fbm{x}_{r}-\fbm{x}_{rd}\\ \sk\left(\fbm{R}_{r}\tbm{R}_{rd}\right)^{\vee}\end{bmatrix}}-w_{t}\tbm{J}_{r}\begin{bmatrix}\wtilbm{x}_{r}-\bbm{R}_{r}\tbbm{R}_{l}\wtilbm{x}_{l}\\ \sk\left(\fbm{R}_{r}\twtilbm{R}_{l}\tbbm{R}_{r}\right)^{\vee}\end{bmatrix}-w_{s}\tbm{J}_{r}\sum^{4}_{i=1}\Big[\fbm{L}^{r}_{li}{}^\mathsf{T}\left(\fbm{s}_{li}-\bbm{s}_{li}\right)+\fbm{L}^{r}_{ri}{}^\mathsf{T}\left(\fbm{s}_{ri}-\bbm{s}_{ri}\right)\Big]
\end{align}
\end{subequations}
\end{figure*}
where $\dotbm{q}_{l}$ and $\dotbm{q}_{r}$ are respectively the negative gradients of the objective function~$V$ with respect to $\fbm{q}_{l}$ and $\fbm{q}_{r}$, and $\sk(\fbm{M})^\vee$ returns the vector associated with the skew-symmetric part of any square matrix~$\fbm{M}$. The next subsection proves the closed-loop system stability.

\subsection{Stability Analysis}

One can investigate the stability of the closed-loop dual-arm manipulation system by the objective function~$V$ defined in \eq{equ14}. Under the joint velocity control inputs~\eqref{equ15}, the kinematics~\eqref{equ1} and \eqref{equ2} of the left and the right manipulators adapt the objective function~$V$ by
\begin{equation}\label{equ16}
\begin{aligned}
\dot{V}={w_{d}\dot{V}_{d}+}w_{t}\dot{V}_{t}+w_{s}\dot{V}_{s}=-\dottbm{q}_{l}\dotbm{q}_{l}-\dottbm{q}_{r}\dotbm{q}_{r}\leq 0\textrm{,}
\end{aligned}
\end{equation}
where the time derivatives of $V_{d}$, $V_{t}$ and $V_{s}$ are presented in detail in \eqref{equ17}, \eqref{equ18} and \eqref{equ19}, respectively. Because the objective function~$V$ is nonnegative and its time derivative is nonpositive, one can thus conclude that the states~$\fbm{q}_l$ and $\fbm{q}_r$ of the system converge to the largest invariant set of
\begin{align*}
\left\{(\fbm{q}_{l},\fbm{q}_{r})\ \big|\ \dotbm{q}_{l}=\dotbm{q}_{r}=\fbm{0}\right\}
\end{align*}
by LaSalle's invariance principle. That is, both arms become stationary asymptotically.

\begin{figure*}
\begin{equation}\label{equ17}
\begin{aligned}
\dot{V}_{d}=&-\frac{1}{2}\sum_{\ast=l,r}\tr\left(\tbm{R}_{\ast d}\dotbm{R}_{\ast}\right)+\sum_{\ast=l,r}(\fbm{x}_{\ast}-\fbm{x}_{\ast d})^\mathsf{T}\dotbm{x}_{\ast}=-\frac{1}{2}\sum_{\ast=l,r}\tr\left(\tbm{R}_{\ast d}\bm{\omega}^{\times}_{\ast}\fbm{R}_{\ast}\right)+\sum_{\ast=l,r}(\fbm{x}_{\ast}-\fbm{x}_{\ast d})^\mathsf{T}\bm{\upsilon}_{\ast}\\
=&-\frac{1}{2}\sum_{\ast=l,r}\tr\left(\bm{\omega}^{\times}_{\ast}\fbm{R}_{\ast}\tbm{R}_{\ast d}\right)+\sum_{\ast=l,r}(\fbm{x}_{\ast}-\fbm{x}_{\ast d})^\mathsf{T}\bm{\upsilon}_{\ast}=\sum_{\ast=l,r}\tfbm{\omega}_{\ast}\sk\left(\fbm{R}_{\ast}\tbm{R}_{\ast d}\right)^{\vee}+(\fbm{x}_{\ast}-\fbm{x}_{\ast d})^\mathsf{T}\bm{\upsilon}_{\ast}\\
=&\sum_{\ast=l,r}\begin{bmatrix}\fbm{x}_{\ast}-\fbm{x}_{\ast d}\\ \sk\left(\fbm{R}_{\ast}\tbm{R}_{\ast d}\right)^{\vee}\end{bmatrix}^\mathsf{T}\fbm{v}_{\ast}=\sum_{\ast=l,r}\Big[\nabla_{\fbm{q}_{\ast}}V_{d}\Big]^\mathsf{T}\dotbm{q}_{\ast}=\begin{bmatrix}\nabla_{\fbm{q}_{l}}V_{d}\\ \nabla_{\fbm{q}_{r}}V_{d}\end{bmatrix}^\mathsf{T}\begin{bmatrix}\dotbm{q}_{l}\\ \dotbm{q}_{r}\end{bmatrix}
\end{aligned}
\end{equation}
\hrulefill
\end{figure*}

\begin{figure*}
\begin{equation}\label{equ18}
\begin{aligned}
\dot{V}_{t}=&-\frac{1}{2}\tr\left(\dotwtilbm{R}_{l}\twtilbm{R}_{r}+\wtilbm{R}_{l}\dottwtilbm{R}_{r}\right)+\left(\tbbm{R}_{l}\wtilbm{x}_{l}-\tbbm{R}_{r}\wtilbm{x}_{r}\right)^\mathsf{T}\left(\tbbm{R}_{l}\dotwtilbm{x}_{l}-\tbbm{R}_{r}\dotwtilbm{x}_{r}\right)\\
=&-\frac{1}{2}\tr\left(\tbbm{R}_{l}\dotbm{R}_{l}\twtilbm{R}_{r}+\wtilbm{R}_{l}\dottbm{R}_{r}\bbm{R}_{r}\right)+\left(\tbbm{R}_{l}\wtilbm{x}_{l}-\tbbm{R}_{r}\wtilbm{x}_{r}\right)^\mathsf{T}\left(\tbbm{R}_{l}\dotbm{x}_{l}-\tbbm{R}_{r}\dotbm{x}_{r}\right)\\
=&-\frac{1}{2}\tr\left(\dotbm{R}_{l}\twtilbm{R}_{r}\tbbm{R}_{l}\right)-\frac{1}{2}\tr\left(\bbm{R}_{r}\wtilbm{R}_{l}\dottbm{R}_{r}\right)+\left(\tbbm{R}_{l}\wtilbm{x}_{l}-\tbbm{R}_{r}\wtilbm{x}_{r}\right)^\mathsf{T}\tbbm{R}_{l}\dotbm{x}_{l}+\left(\tbbm{R}_{r}\wtilbm{x}_{r}-\tbbm{R}_{l}\wtilbm{x}_{l}\right)^\mathsf{T}\tbbm{R}_{r}\dotbm{x}_{r}\\
=&-\frac{1}{2}\tr\left(\bm{\omega}^{\times}_{l}\fbm{R}_{l}\twtilbm{R}_{r}\tbbm{R}_{l}\right)+\frac{1}{2}\tr\left(\bbm{R}_{r}\wtilbm{R}_{l}\tbm{R}_{r}\bm{\omega}^{\times}_{r}\right)+\left(\tbbm{R}_{l}\wtilbm{x}_{l}-\tbbm{R}_{r}\wtilbm{x}_{r}\right)^\mathsf{T}\tbbm{R}_{l}\bm{\upsilon}_{l}+\left(\tbbm{R}_{r}\wtilbm{x}_{r}-\tbbm{R}_{l}\wtilbm{x}_{l}\right)^\mathsf{T}\tbbm{R}_{r}\bm{\upsilon}_{r}\\
=&\tfbm{\omega}_{l}\sk\left(\fbm{R}_{l}\twtilbm{R}_{r}\tbbm{R}_{l}\right)^{\vee}+\tfbm{\omega}_{r}\sk\left(\fbm{R}_{r}\twtilbm{R}_{l}\tbbm{R}_{r}\right)^{\vee}+\left(\wtilbm{x}_{l}-\bbm{R}_{l}\tbbm{R}_{r}\wtilbm{x}_{r}\right)^\mathsf{T}\bm{\upsilon}_{l}+\left(\wtilbm{x}_{r}-\bbm{R}_{r}\tbbm{R}_{l}\wtilbm{x}_{l}\right)^\mathsf{T}\bm{\upsilon}_{r}\\
=&\begin{bmatrix}\wtilbm{x}_{l}-\bbm{R}_{l}\tbbm{R}_{r}\wtilbm{x}_{r}\\ \sk\left(\fbm{R}_{l}\twtilbm{R}_{r}\tbbm{R}_{l}\right)^{\vee}\end{bmatrix}^\mathsf{T}\fbm{v}_{l}+\begin{bmatrix}\wtilbm{x}_{r}-\bbm{R}_{r}\tbbm{R}_{l}\wtilbm{x}_{l}\\ \sk\left(\fbm{R}_{r}\twtilbm{R}_{l}\tbbm{R}_{r}\right)^{\vee}\end{bmatrix}^\mathsf{T}\fbm{v}_{r}=\begin{bmatrix}\nabla_{\fbm{q}_{l}}V_{t}\\ \nabla_{\fbm{q}_{r}}V_{t}\end{bmatrix}^\mathsf{T}\begin{bmatrix}\dotbm{q}_{l}\\ \dotbm{q}_{r}\end{bmatrix}
\end{aligned}
\end{equation}
\hrulefill
\end{figure*}

\begin{figure*}
\begin{equation}\label{equ19}
\begin{aligned}
\dot{V}_{s}=&\sum^{4}_{i=1}\begin{bmatrix}\fbm{s}_{li}-\bbm{s}_{li}\\ \fbm{s}_{ri}-\bbm{s}_{ri}\end{bmatrix}^\mathsf{T}\begin{bmatrix}\dotbm{s}_{li}\\ \dotbm{s}_{ri}\end{bmatrix}=\sum^{4}_{i=1}\begin{bmatrix}\fbm{s}_{li}-\bbm{s}_{li}\\ \fbm{s}_{ri}-\bbm{s}_{ri}\end{bmatrix}^\mathsf{T}\begin{bmatrix}\fbm{L}^{l}_{li} &\fbm{L}^{r}_{li}\\ \fbm{L}^{l}_{ri} &\fbm{L}^{r}_{ri}\end{bmatrix}\begin{bmatrix}\fbm{v}_{l}\\ \fbm{v}_{r}\end{bmatrix}=\sum^{4}_{i=1}\left(\begin{bmatrix}\fbm{L}^{l}_{li} &\fbm{L}^{r}_{li}\\ \fbm{L}^{l}_{ri} &\fbm{L}^{r}_{ri}\end{bmatrix}^\mathsf{T}\begin{bmatrix}\fbm{s}_{li}-\bbm{s}_{li}\\ \fbm{s}_{ri}-\bbm{s}_{ri}\end{bmatrix}\right)^\mathsf{T}\begin{bmatrix}\fbm{v}_{l}\\ \fbm{v}_{r}\end{bmatrix}\\
=&\sum^{4}_{i=1}\begin{bmatrix}\fbm{L}^{l}_{li}{}^\mathsf{T}\left(\fbm{s}_{li}-\bbm{s}_{li}\right)+\fbm{L}^{l}_{ri}{}^\mathsf{T}\left(\fbm{s}_{ri}-\bbm{s}_{ri}\right)\\ \fbm{L}^{r}_{li}{}^\mathsf{T}\left(\fbm{s}_{li}-\bbm{s}_{li}\right)+\fbm{L}^{r}_{ri}{}^\mathsf{T}\left(\fbm{s}_{ri}-\bbm{s}_{ri}\right)\end{bmatrix}^\mathsf{T}\begin{bmatrix}\fbm{v}_{l}\\ \fbm{v}_{r}\end{bmatrix}=\begin{bmatrix}\nabla_{\fbm{q}_{l}}V_{s}\\ \nabla_{\fbm{q}_{r}}V_{s}\end{bmatrix}^\mathsf{T}\begin{bmatrix}\dotbm{q}_{l}\\ \dotbm{q}_{r}\end{bmatrix}
\end{aligned}
\end{equation}
\hrulefill
\end{figure*}

From \eqs{equ17}{equ19}, one can find that the joint velocity control inputs~$\dotbm{q}_{l}$ and $\dotbm{q}_{r}$ of the left and the right arms designed in \eqref{equ15} can be written as
\begin{subequations}\label{equ20}
\begin{align}
\dotbm{q}_{l}=&-w_{d}\nabla_{\fbm{q}_{l}}V_{d}-w_{t}\nabla_{\fbm{q}_{l}}V_{t}-w_{s}\nabla_{\fbm{q}_{l}}V_{s}\textrm{,}\label{equ20a}\\
\dotbm{q}_{r}=&-w_{d}\nabla_{\fbm{q}_{r}}V_{d}-w_{t}\nabla_{\fbm{q}_{r}}V_{t}-w_{s}\nabla_{\fbm{q}_{r}}V_{s}\textrm{,}\label{equ20b}
\end{align}
\end{subequations}
where $\nabla_{\fbm{q}_{\ast}}V_{d}$, $\nabla_{\fbm{q}_{\ast}}V_{t}$ and $\nabla_{\fbm{q}_{\ast}}V_{s}$ are the gradients of the objective functions~$V_{d}$, $V_{t}$ and $V_{s}$ with respect to $\fbm{q}_{\ast}$ for $\ast=l,r$. Then, the fact that $\dotbm{q}_{\ast}\to\fbm{0}$ together with \eqref{equ20} imply that the gradient of the objective function~$V$ tends to zero, i.e.,
\begin{align*}
\begin{bmatrix}\dotbm{q}_{l}\\ \dotbm{q}_{r}\end{bmatrix}=-\begin{bmatrix}\nabla_{\fbm{q}_{l}}V\\ \nabla_{\fbm{q}_{r}}V\end{bmatrix}=-\nabla V\to 0
\end{align*}
as time goes to infinity, where $\nabla_{\fbm{q}_{\ast}}V$ are the gradients of the objective function~$V$ with respect to $\fbm{q}_{\ast}$. One can thus conclude that the objective function~$V$ converges to its minimum.

Note that the objective function~$V$ is not necessarily globally strictly convex with respect to $\fbm{q}_{l}$ and $\fbm{q}_{r}$. The steady-state configuration of the dual-arm system can possibly be a local minimum point of $V$. Nevertheless, $V=0$ at the initial time is exactly its global minimum by its definition, which implies that the proposed controls~\eqref{equ15} reduce $V$ back to zero asymptotically once it diverges a bit from zero throughout the cooperative manipulation. Therefore, both the pose synchronization control~\eqref{equ5} and the enhanced cooperation control~\eqref{equ10} objectives are fulfilled.

\section{Discussion}\label{sec: discussion}

This section discusses the state and input constraints of manipulators and visual servoing issues.

\subsection{State and Input Constraints}

Implementing the proposed control on a real-robot platform needs to cope with the system's state and input constraints. For the velocity-controlled manipulators~$\ast=l,r$, their joint angles~$\fbm{q}_{\ast}$ and velocities~$\dotbm{q}_{\ast}$ are the states and inputs of the system, respectively. Typical state constraints of the velocity-control manipulators are imposed to avoid the mechanical limits of their joints, obstacles in the surrounding environment, and collisions between their bodies. Together, all these state constraints can be unified into an inequality of the manipulators' joint angles as follows
\begin{align*}
h(\fbm{q}_{l},\fbm{q}_{r})\leq 0\textrm{.}
\end{align*}
The input constraints are imposed to accommodate the limits of the left~$\ast=l$ and the right~$\ast=r$ arms' joint velocities. Namely, the joint velocity control inputs~$\dotbm{q}_{\ast}$ cannot exceed their maximum~$\dotbm{q}^{\max}_{\ast}$ and the minimum~$\dotbm{q}^{\min}_{\ast}$ values as follows
\begin{align*}
\dotbm{q}^{\min}_{\ast}\leq \dotbm{q}_{\ast}\leq \dotbm{q}^{\max}_{\ast}\textrm{.}
\end{align*}

Under the state constraints, the joint velocity control inputs to the manipulators can be designed by solving the optimization problem~\eqref{equ14} with the penalty function
\begin{align*}
\psi(\fbm{q}_{l},\fbm{q}_{r})=\exp\left[\frac{\overline{h}-h(\fbm{q}_{l},\fbm{q}_{r})}{h(\fbm{q}_{l},\fbm{q}_{r})}\right]-\frac{\overline{h}-h(\fbm{q}_{l},\fbm{q}_{r})}{h(\fbm{q}_{l},\fbm{q}_{r})}-1
\end{align*}
for $\overline{h}<h(\fbm{q}_{l},\fbm{q}_{r})<0$ and $\psi(\fbm{q}_{l},\fbm{q}_{r})=0$ for $h(\fbm{q}_{l},\fbm{q}_{r})\leq\overline{h}$ with $\overline{h}<0$. Then, the input constraints are guaranteed by scaling the velocity inputs~$\dotbm{q}_{\ast}$ with factors
\begin{align*}
\alpha_{\ast}=\max\limits_{i=1,\cdots,6}\left(\frac{\dot{q}_{\ast i}}{\dot{q}^{\max}_{\ast i}},\frac{\dot{q}_{\ast i}}{\dot{q}^{\min}_{\ast i}}, 1\right)\textrm{,}
\end{align*}
where $\dot{q}_{\ast i}$, $\dot{q}^{\max}_{\ast i}$ and $\dot{q}^{\min}_{\ast i}$ are the $i$-th elements of $\dotbm{q}_{\ast}$, $\dotbm{q}^{\max}_{\ast}$ and $\dotbm{q}^{\min}_{\ast}$, respectively. Together, the left~$\ast=l$ and the right~$\ast=r$ arms cooperatively solve the optimization problem~\eqref{equ14} under the state and input constraints by their joint velocity inputs
\begin{align*}
\dotbm{q}_{\ast}=-\frac{1}{\alpha_{\ast}}\Big[\nabla_{\fbm{q}_{\ast}}V+\nabla_{\fbm{q}_{\ast}}\psi(\fbm{q}_{l},\fbm{q}_{r})\Big]\textrm{,}
\end{align*}
where $\nabla_{\fbm{q}_{\ast}}\psi(\fbm{q}_{l},\fbm{q}_{r})$ are the gradients of the penalty function~$\psi(\fbm{q}_{l},\fbm{q}_{r})$ with respect to $\fbm{q}_{\ast}$.

\subsection{Visual Servoing Issues}

The system performance is limited by the low data rate of vision sensors and the latency of feature extracting algorithms. Thus, extended Kalman filters have been integrated to predict and track moving targets~\cite{Oliva2022ICRA}, and sampled-data controls have been developed to stabilize the pick and place operations~\cite{Constanzo2024TCST}. The lighting conditions are also unignorable for deploying vision-based controls in complex environments. Poor illumination conditions significantly degrade the performance of vision sensors. Thus, the mutual information~\cite{Marchand2011TRO} and the dense depth map~\cite{Marchand2014TRO} have robustified visual servoing against illumination variations at the cost of positioning accuracy. Event-based cameras are promising for overcoming the low data rate and the sensitivity to illumination of traditional vision sensors.

Large objects can possibly occlude markers mounted on the two manipulators. The visual servoing control then becomes invalid. In this situation, one can attach fiducial markers to the object and design visual servoing algorithms for the two manipulators with wrist-mounted cameras to precisely regulate their poses relative to the markers, thereby securely holding and manipulating the objects. However, the kinematic and dynamic models of the closed-loop system would become entirely different due to the possible relative motions between robots and objects. Future research will design new control algorithms to handle the manipulation of large objects.

PBVS and 2-1/2-D visual servoing~\cite{Malis1999TRA} both possess larger domains of attraction than IBVS by adding some 3D information to the visual features. Yet, estimating the full or partial poses of markers/cameras requires more complicated image processing algorithms, which may induce extra latencies and degrade the system's responsiveness. Also, pose estimation algorithms are vulnerable to sensing noises. Even small errors in the image measurements can cause significant errors in the estimated pose, thereby undermining the system's positioning accuracy. Developing fast and robust pose estimation algorithms remains a future work.
\color{black}

\section{Experimental Results}\label{sec: experimental results}

\begin{figure}[!hb]
\centering
\begin{overpic}[width=.95\columnwidth]{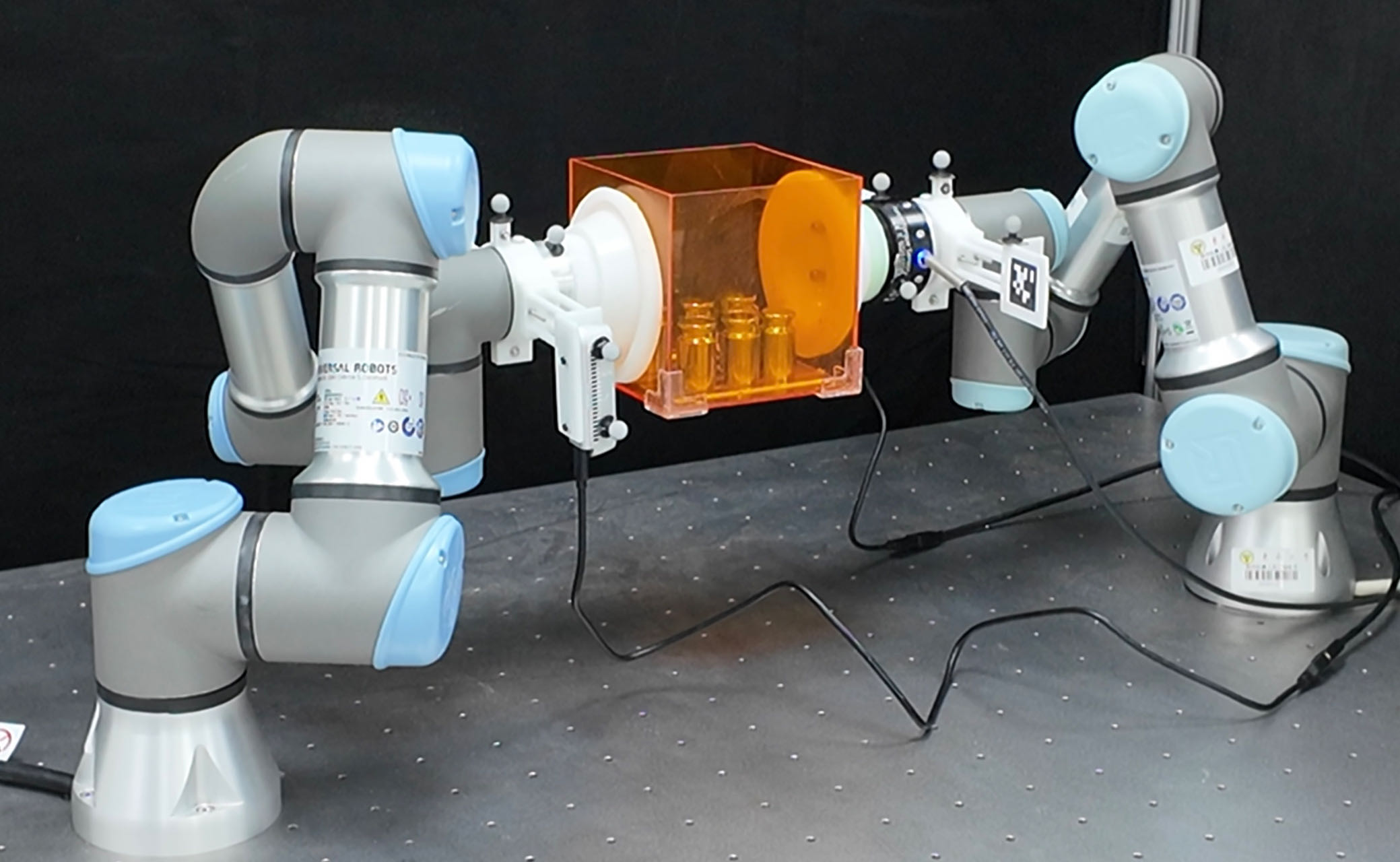}
\linethickness{1pt}
\put(60,12){\color{red}\vector(-1,0.7){7}}
\put(60,12){\color{green}\vector(-1,-0.2){10}}
\put(60,12){\color{blue}\vector(0,1){10}}
\end{overpic}
\caption{The experimental setup includes two Universal Robots UR3/UR3e manipulators. Both arms equip elastic tools on their flanges to transport a cubic box via frictional contacts. The camera mounted on the wrist of each arm captures the marker mounted on the wrist of the other arm.}
\label{fig3}
\end{figure}

This section evaluates the proposed control via comparative experiments on a real-robot platform. The experimental setup includes a pair of Universal Robots UR3/UR3e manipulators, as shown in \fig{fig3}. The task of the dual-arm system is to {hold} the target object, a $15$~cm$\times 15$~cm$\times 15$~cm cubic orange box of weight {$2.28$~kg}, without fixtures and transport it from the initial pose to the target pose while passing $3$ waypoints. Two customized bowl-shaped grippers with silicone covers {of shore 0A hardness and $25$~mm thickness} are mounted to the flanges of both arms to prevent them from over-squeezing the manipulated object. {The interaction wrenches between the end effectors of the two arms and the box are measured by the built-in force/torque sensor of the UR3e arm.} On the wrist of each arm, an Intel RealSense D435 camera captures the four corner points of the AprilTag marker from the 36h11 family mounted on the wrist of the other arm at about $30$~Hz. The coordinates of the corner points on the image planes of the cameras serve as the image features for IBVS. The controls of the UR3/UR3e manipulators at $125$~Hz/$500$~Hz are implemented in C++ programs that run on a single Ubuntu machine. {The Real-Time Data Exchange (RTDE) synchronizes control programs with the UR robot hardware over a standard TCP/IP connection without breaking real-time properties.}

In all experiments, the left and right manipulators locate their bases at $(0,-0.55,0)$~m and $(0,0.55,0)$~m and start their joint coordinates with $(-30^\circ,-140^\circ,-85^\circ,45^\circ,75^\circ,0^\circ)$. By the Monte Carlo algorithm, \tab{table1} lists the maximum and minimum coordinates that the end effectors of the two arms can reach. Accounting for the limited workspaces, the two arms transport the object to move along $3$ paths. Each path has two waypoints~$B_o$ and $C_o$ other than the initial/final pose~$A_o$, of which the positions and orientations are listed in \tab{table2}.

\begin{table}
\centering
\caption{The left and the right arms' workspaces in meters.}\label{table1}
\renewcommand{\arraystretch}{1.5}
\begin{tabular}{|c|c|c|c|c|}
\hlineB{3}
\multicolumn{2}{|c|}{arms}& $x$-coordinate & $y$-coordinate & $z$-coordinate \\
\hlineB{3}
\multirow{2}{*}{left}  & min & $-0.4375$ & $-0.0083$ & $-0.3497$ \\
                       & max & $+0.4394$ & $+0.5788$ & $+0.6389$ \\
\hline
\multirow{2}{*}{right} & min & $-0.4394$ & $+0.1289$ & $-0.3497$ \\
                       & max & $+0.4375$ & $+0.7408$ & $+0.6389$ \\
\hlineB{3}
\end{tabular}
\end{table}

\begin{table*}
\centering
\caption{The position coordinates and the roll-pitch-yaw angles of the object's waypoints on $3$ paths.}\label{table2}
\renewcommand{\arraystretch}{1.5}
\begin{tabular}{|c|c|c|c|c|}
\hlineB{3}
\multicolumn{2}{|c|}{paths}                 & path1 & path2 & path3 \\
\hlineB{3}
\multirow{2}{*}{waypoint~$B_o$}    &position [m]     & $(0.13,-0.03,-0.02)$   & $(0,-0.1,-0.05)$  & $(-0.1,-0.05,0)$ \\
                              &orientation [rad]  & $(\pi/10,-\pi/18,0)$   & $(0,-\pi/18,0)$   & $(-\pi/12,-\pi/12,\pi/12)$ \\
\hline
\multirow{2}{*}{waypoint~$C_o$}    &position [m]     & $(-0.03,0.01,0)$       & $(-0.1,0.05,0.05)$   & $(0.1,-0.05,0)$ \\
                              &orientation [rad]  & $(-\pi/12,\pi/12,0)$   & $(\pi/18,0,0)$   & $(-\pi/12,\pi/12,-\pi/10)$ \\
\hlineB{3}
\end{tabular}
\end{table*}

\begin{figure*}
\centering
\subfigure[Path1.]{
    \label{fig4a}
    \begin{overpic}[width=0.625\columnwidth]{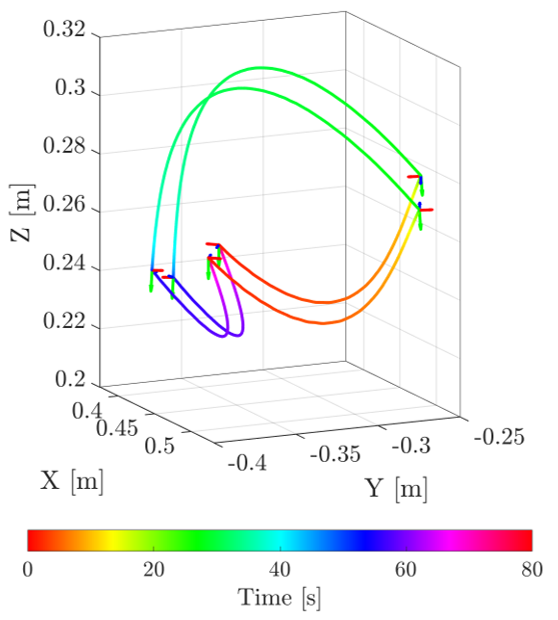}
    \put(35,62){\footnotesize\textcolor{black}{$A_\ast$}}
    \put(70,67){\footnotesize\textcolor{black}{$B_\ast$}}
    \put(22,48){\footnotesize\textcolor{black}{$C_\ast$}}
	\end{overpic}
}
\subfigure[Path2.]{
    \label{fig4b}
    \begin{overpic}[width=0.625\columnwidth]{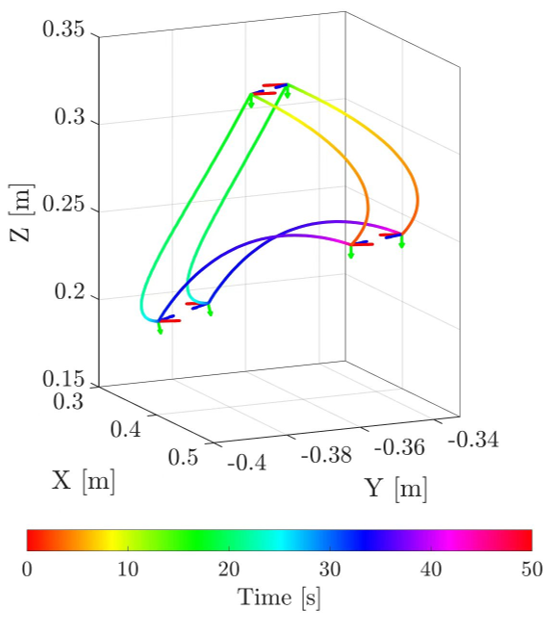}
    \put(59,57){\footnotesize\textcolor{black}{$A_\ast$}}
    \put(38,88){\footnotesize\textcolor{black}{$B_\ast$}}
    \put(30,45){\footnotesize\textcolor{black}{$C_\ast$}}
	\end{overpic}
}
\subfigure[Path3.]{
    \label{fig4c}
    \begin{overpic}[width=0.625\columnwidth]{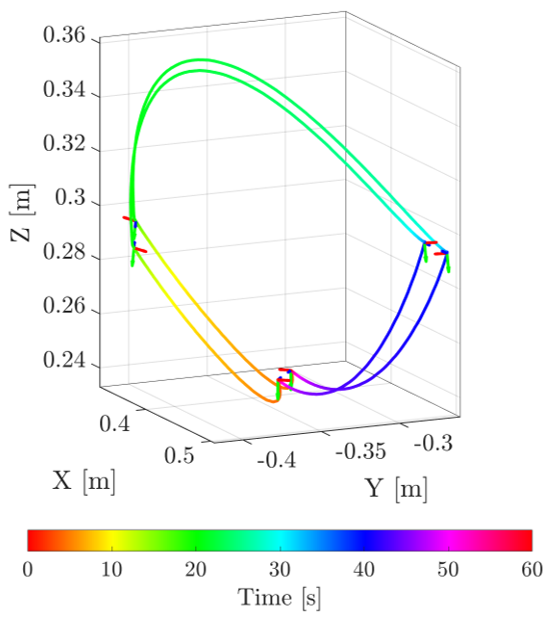}
    \put(42,42){\footnotesize\textcolor{black}{$A_\ast$}}
    \put(24,62){\footnotesize\textcolor{black}{$B_\ast$}}
    \put(69,62){\footnotesize\textcolor{black}{$C_\ast$}}
	\end{overpic}
}
\caption{The Cartesian paths and the waypoints' orientations of the two arms' end effectors when the object moves along path1, path2 and path3, respectively. The final poses of both arms coincide with their initial poses.}
\label{fig4}
\end{figure*}

\fig{fig4} plots the positions and the orientations of the two arms' end effectors when the object moves along path1, path2 and path3, respectively. The colors of the paths change with time as indicated in the color bar, and the coordinate frames at the waypoints mark the end-effector orientations of both arms. The left~$\ast=l$ and the right~$\ast=r$ arms start with their initial poses at the waypoints~$A_\ast$, move towards the waypoints~$B_\ast$ and $C_\ast$, and back to the waypoints~$A_\ast$. Here, the poses of the waypoints are exactly the desired poses~$(\fbm{R}_{\ast d}, \fbm{x}_{\ast d})\in SE(3)$ in the controls~\eqref{equ15} of the two arms.

\begin{figure}
    \centering
    \includegraphics[width=\columnwidth]{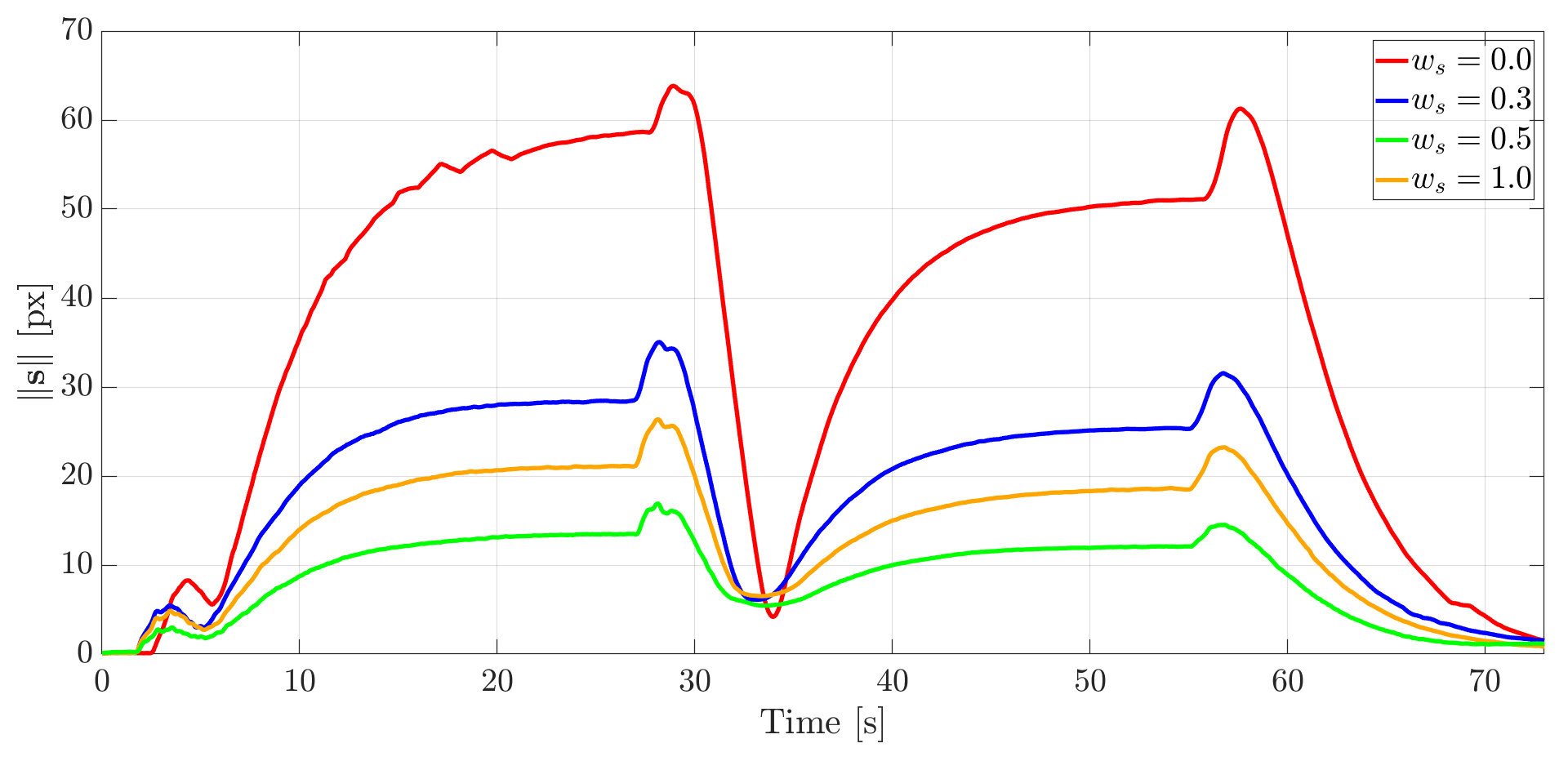}
    \caption{The errors of the image features~$\fbm{s}=(\tbm{s}_l,\tbm{s}_r)^\mathsf{T}$ from their desired (initial) values when transporting the object along path1 during the first group of experiments with $w_d=1.0$ and $w_t=1-w_s$.}
    \label{fig5}
\end{figure}

Two groups of experiments examine the performances of the dual-arm system under the proposed control by picking different sets of gains~$w_d$, $w_t$ and $w_s$. With $w_d=1.0$ and $w_d=1.5$, the first and the second groups transport the object at a relatively lower and higher speed, respectively. In each group, the dual-arm system performs four experimental trials via setting $w_s=0.0$, $w_s=0.3$, $w_s=0.5$ and $w_s=1.0$ in ascending order while letting $w_t=1-w_s$. Basically, $w_s$ weighs the impact of the IBVS controls on the coordination of the two arms. When $w_s=0.0$ and $w_s=1.0$, the proposed control becomes a purely pose synchronization control~\cite{Spong2012TAC} and a purely IBVS control, respectively.

\begin{figure}
    \centering
    \includegraphics[width=1\columnwidth]{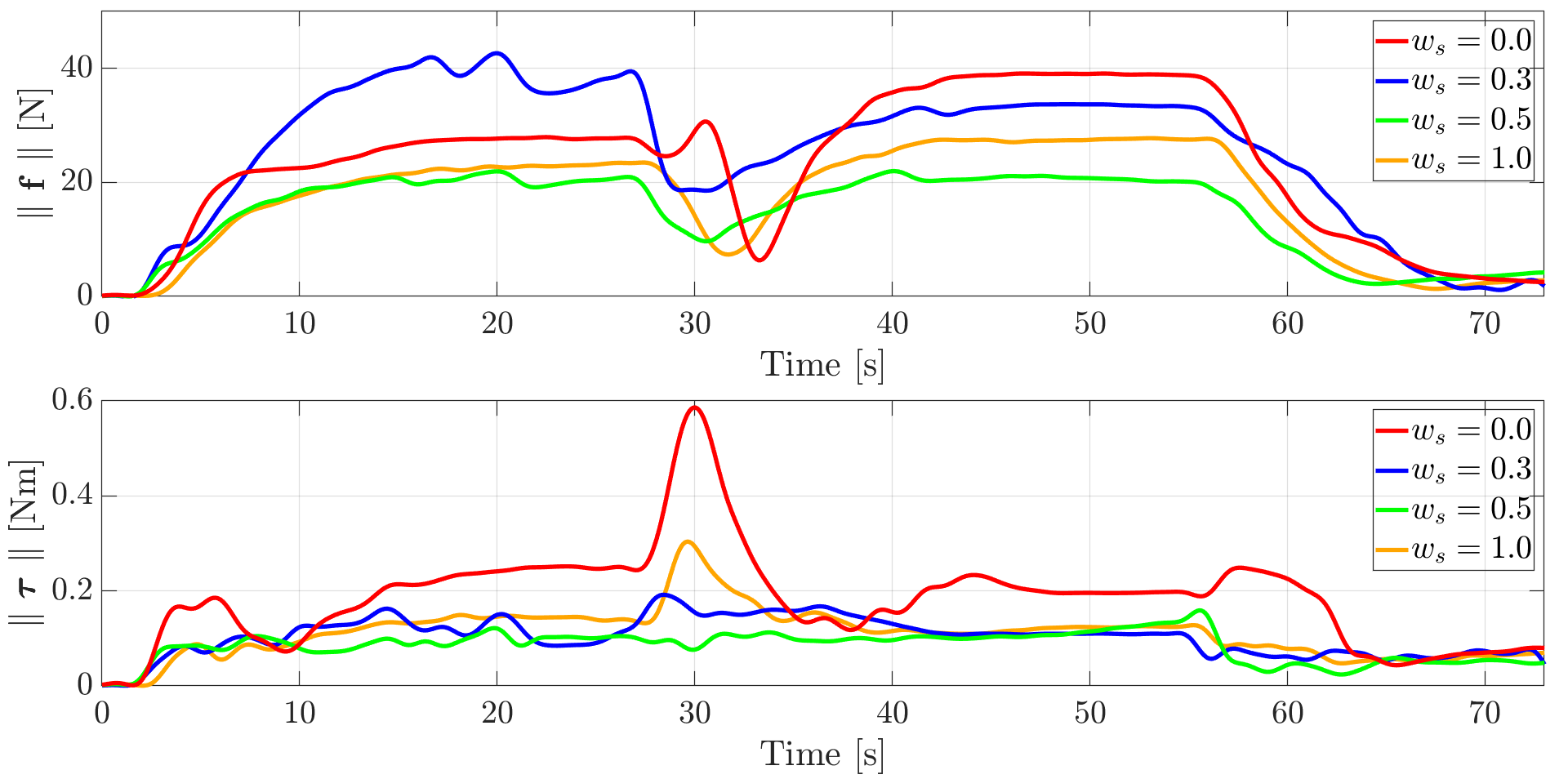}
    \caption{The deviations of the interaction forces (the top subplot) and torques (the bottom subplot) between the end effector of the UR3e manipulator and the object when transporting the object along path1 during the first group of experiments with $w_d=1.0$ and $w_t=1-w_s$.}
    \label{fig6}
\end{figure}

At a relatively lower speed, the first group of experiments all successfully transport the object along the planned three paths. Nevertheless, the performances of the dual-arm manipulation system change with the control gains~$w_s$ and $w_t=1-w_s$. \fig{fig5} depicts the errors of the image features from their desired values when the two arms transport the object to move along path1. The gain~$w_s$ grows in the order of $0.0$, $0.3$, $0.5$ and $1.0$. For most of the time, the purely pose synchronization control ($w_s=0.0$) exhibits the greatest error of image features. Its maximum error is $2$ times larger than other trials. As $w_s$ raises to $0.3$ and $0.5$ sequentially, the errors of image features reduce accordingly. When the control becomes purely IBVS ($w_s=1.0$), the error grows again, staying between the errors of the trials~$w_s=0.3$ and $w_s=0.5$. Overall, setting equal weights~$w_s=w_t=0.5$ for the pose synchronization and IBVS controls obtains the smallest error. The error profiles display a similar trend in all trials, which suggests that they are related to the manipulators' postures as well.

\fig{fig6} presents the deviations of the interaction forces and torques between the end effector of the UR3e arm and the object. Corresponding to \fig{fig5}, it delivers that the interaction forces and torques vary with the image feature errors throughout the object transportation. Roughly speaking, the interaction forces and torques increase/decrease with the errors of image features. From the $7$~s to the $27$~s, however, the interaction forces deviate less when the pose synchronization control functions alone than together with the IBVS control using $w_s=0.3$. One can also notice that, at the $30$~s, the torque deviations of the purely pose synchronization control and the purely IBVS control both reach their peaks. By the contrast, weighing the pose synchronization control and the IBVS control equally with $w_s=w_t=0.5$ maintains both the force and torque deviations flattest and smallest among all the four trials.

\begin{figure}
\centering
\subfigure[$\fbm{s}_l$: $w_s=0.0$.]{
    \label{fig7a}
    \begin{overpic}[height=2.7cm]{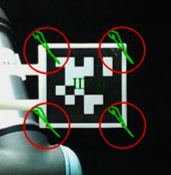}
	\end{overpic}
}
\subfigure[$\fbm{s}_l$: $w_s=1.0$.]{
    \label{fig7b}
    \begin{overpic}[height=2.7cm]{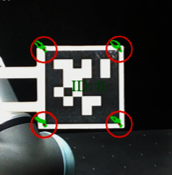}
	\end{overpic}
}
\subfigure[$\fbm{s}_l$: $w_s=0.5$.]{
    \label{fig7c}
    \begin{overpic}[height=2.7cm]{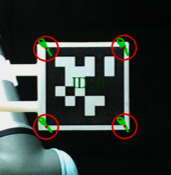}
	\end{overpic}
}

\subfigure[$\fbm{s}_r$: $w_s=0.0$.]{
    \label{fig7d}
    \begin{overpic}[height=2.7cm]{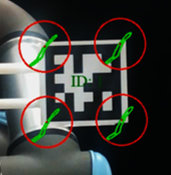}
	\end{overpic}
}
\subfigure[$\fbm{s}_r$: $w_s=1.0$.]{
    \label{fig7e}
    \begin{overpic}[height=2.7cm]{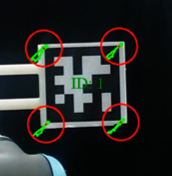}
	\end{overpic}
}
\subfigure[$\fbm{s}_r$: $w_s=0.5$.]{
    \label{fig7f}
    \begin{overpic}[height=2.7cm]{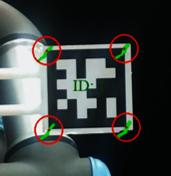}
	\end{overpic}
}
\caption{The trajectories~$\fbm{s}_\ast$ of the markers' corners on the images of the left~$\ast=l$ and the right~$\ast=r$ cameras when transporting the object along path3 in the second group of experiments with $w_d=1.5$ and $w_t=1-w_s$. The green curves are the image trajectories. The centers and radii of red circles indicate the original locations and the maximum deviations of the markers' corners on the images, respectively.}
\label{fig7}
\end{figure}

At a relatively higher speed, the second group of experiments encounter some failures. In particular, both the purely pose synchronization control ($w_s=0.0$) and the purely IBVS control ($w_s=1.0$) cause the object to fall down or slide when transporting it along path1 and path3. In contrast, weighing the pose synchronization control and the IBVS control equally with $w_s=w_t=0.5$ succeeds in transporting the object along all three paths. Besides the accompanying videos, \fig{fig7} compares the image trajectories of the markers' corners when the two arms transport the object along path3 with $w_s=0.0$, $w_s=1.0$ and $w_s=0.5$, respectively. One can find that the markers' corners exhibit the greatest deviations with $w_s=0.0$, the moderate deviations with $w_s=1.0$, and the least deviations with $w_s=0.5$. Correspondingly, the object falls down, slides, and holds on in the respective experimental trials.
\color{black}

\section{Conclusion}\label{sec: conclusion}

This paper has developed an IBVS approach for enhancing the cooperation of dual-arm manipulation. By coordinating the two arms with their kinematics and joint angle measurements, the conventional synchronization control leads to significant end-effector pose errors of the dual-arm manipulation system when systematic errors exist. In contrast, the proposed control can mitigate the pose synchronization errors because the IBVS enhancement is robust to the kinematic errors of manipulators. Technically, the paper has modelled the kinematics of a dual-arm manipulation system, has tailored the interaction matrix to the dual-arm system, has designed the joint velocity control inputs with rigorous stability proof, and has discussed the state and input constraints of manipulators and visual servoing issues. Future studies will control the interaction forces between manipulators and objects and will coordinate them to assemble parts and components on floating bases.

\bibliography{IEEEabrv,reference}

\end{document}